\documentclass[11pt]{article}

\usepackage[preprint]{acl}
\usepackage{times}
\usepackage{latexsym}
\usepackage[T1]{fontenc}
\usepackage[utf8]{inputenc}
\usepackage{microtype}
\usepackage{inconsolata}

\usepackage{amsmath,amsfonts,bm}

\def\eqref#1{equation~\ref{#1}}

\def\1{\bm{1}}

\DeclareMathAlphabet{\mathsfit}{\encodingdefault}{\sfdefault}{m}{sl}
\SetMathAlphabet{\mathsfit}{bold}{\encodingdefault}{\sfdefault}{bx}{n}

\usepackage{booktabs}
\usepackage{multirow}
\usepackage{graphicx}
\usepackage{xcolor}
\usepackage[breakable]{tcolorbox}
\usepackage{colortbl}
\usepackage{float}
\usepackage{algorithm}
\usepackage{algpseudocode}
\usepackage{hyperref}
\usepackage{url}

\newtcolorbox{promptbox}[2]{
    breakable, colback=#1!4, colframe=#1!40, arc=4pt,
    title={#2}, coltitle=black, fonttitle=\bfseries\small,
    fontupper=\small\ttfamily, halign=flush left,
    before upper={\setlength{\parindent}{0pt}\noindent},
    left=6pt, right=6pt, top=4pt, bottom=4pt
}

\title{VASPO: Self-Distillation with Privileged Value Estimation}

\author{
Fei Wu\textsuperscript{1}, Zhenrong Zhang\textsuperscript{2},
Jianshu Zhang\textsuperscript{2}, Jun Du\textsuperscript{1}\thanks{Corresponding author.}\\
\textsuperscript{1}University of Science and Technology of China\\
\textsuperscript{2}IFLYTEK Research
}
\hypersetup{pdftitle={VASPO: Self-Distillation with Privileged Value Estimation},
pdfauthor={Fei Wu, Zhenrong Zhang, Jianshu Zhang, Jun Du},pdfsubject={},pdfkeywords={}}

\begin{document}

\maketitle

\begin{abstract}
On-policy self-distillation provides dense supervision for reasoning models via a privileged self-teacher. However, this supervision reflects local token preferences rather than the expected return of an unguided student continuation. Recent methods combine this signal with GRPO. They use GRPO for update directions and the teacher for signal scaling. However, we demonstrate a more effective integration within PPO. Specifically, we identify complementary positional dynamics: privileged information enhances a critic's value-prediction accuracy toward the response's end, whereas self-teacher guidance is strong initially, drops mid-generation, and rebounds finally. These insights motivate \emph{Value-Augmented Self-Distillation Policy Optimization} (VASPO), wherein a self-teacher and critic evaluate identical privileged information at student-generated prefixes. The critic computes token-level PPO advantages from predicted continuation returns, while the self-teacher modulates selected positive advantages via bounded probability ratios. Across six mathematical reasoning benchmarks, VASPO improves mean accuracy over PPO by 2.59 and 1.81 percentage points on Qwen3-4B and OLMo3-7B variants, respectively, and achieves the highest mean accuracy among evaluated methods on Knights and Knaves. Our code is available at \url{https://github.com/cii030/VASPO}.
\end{abstract}

\begin{figure}[t]
    \centering
    \includegraphics[width=\columnwidth]{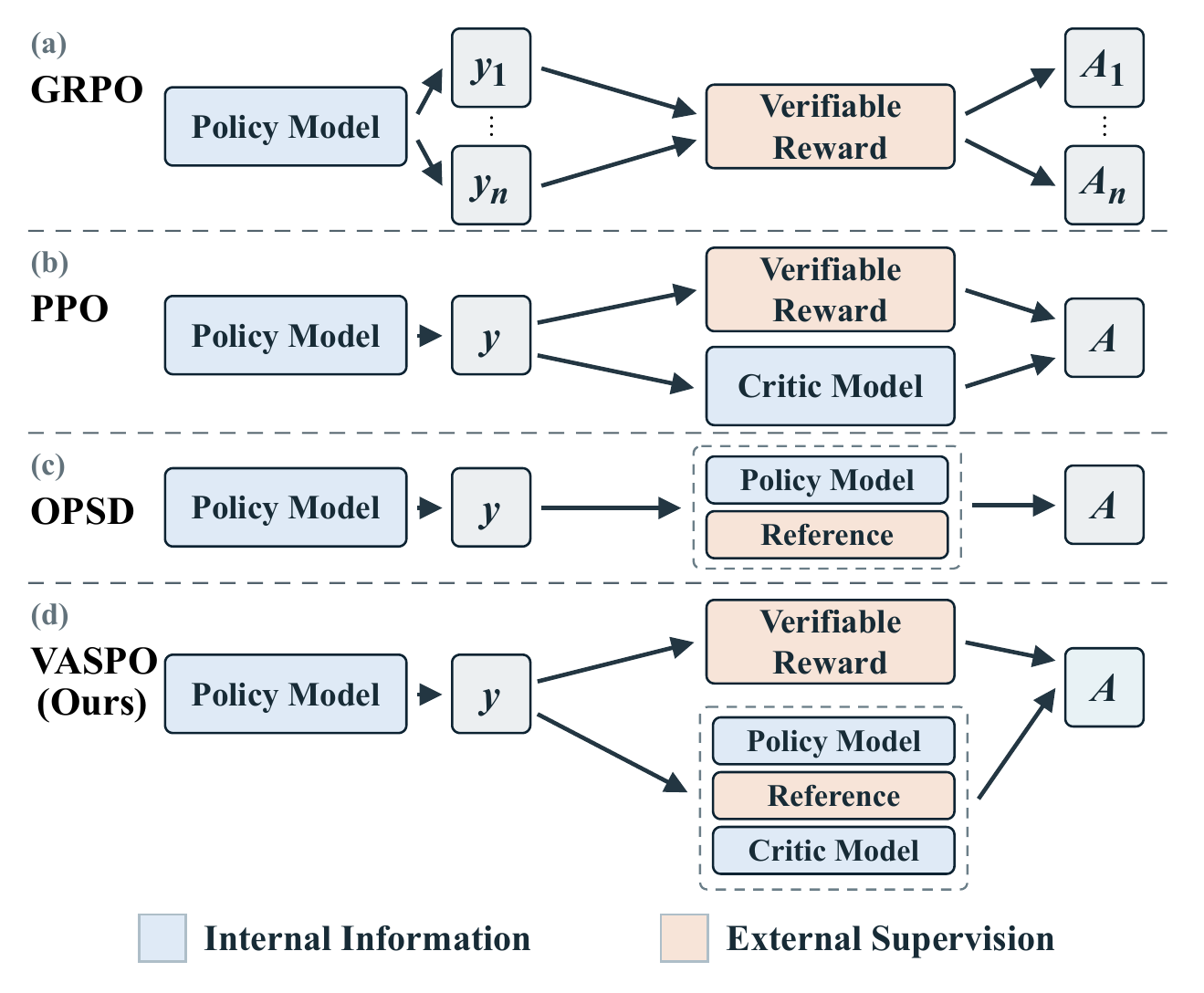}
    \caption{\textbf{Training signals in GRPO, PPO, OPSD, and VASPO.}
    VASPO shares privileged information for token guidance and value
    estimation.}
    \label{fig:teaser}
\end{figure}

\section{Introduction}
\label{sec:introduction}

Training reasoning models from verified final answers requires assigning
credit to intermediate decisions. Reinforcement learning with verifiable
rewards (RLVR) supplies a reliable outcome but does not identify which
tokens contributed to it \citep{guo2025deepseek,xiao2026mimo}. Group Relative
Policy Optimization (GRPO) compares several responses to the same problem
and broadcasts each response's advantage across its tokens \citep{grpo}.
Proximal Policy Optimization (PPO) instead learns a value model, or critic,
to turn the outcome into advantages
\citep{schulman2017proximalpolicyoptimizationalgorithmsppo}. Both approaches
must learn from a reward that arrives only after the response is complete.

On-policy self-distillation (OPSD) supplies a denser, token-level signal at
student-generated prefixes. The self-teacher reads privileged
information, such as a worked solution or a ground-truth answer, and scores the sampled tokens
\citep{zhao2026opsd}. This guidance can distinguish tokens within a response,
but its preferences conditioned on privileged information are not
verifier-grounded measures of the student's eventual success. Privileged
context can shift the signal toward style rather than task content
\citep{pan2026rlcsdreinforcementlearningcontrastive}, suppress deliberation in
reasoning models \citep{kaur2026rethinking}, or yield unreliable targets at
some prefixes \citep{liu2026teachertokens}. These observations motivate
combining token guidance with outcome feedback while measuring when that
guidance is useful.

Recent combinations use verifier outcomes to anchor or route privileged
teacher guidance. RLSD modulates group-relative advantages with
teacher--student signals
\citep{yang2026selfdistilledrlvr};
SSPO applies step-level weights to GRPO in deep search
\citep{wu2026step}. SRPO routes correct and failed responses between GRPO
and self-distillation \citep{li2026unifying}. These methods use privileged
information in the teacher branch. None uses the same privileged information
to train a PPO-style prefix-value estimator for clean student returns. This
leaves a separate use of that information unexplored: estimating the student's
continuation return before assigning token credit.

Prior work has demonstrated the effectiveness of providing reference answers as input to the critic \citep{zhang2026redippo}.
Teacher preferences and continuation values serve fundamentally distinct roles at the same student prefix. While the teacher evaluates immediate token choices using privileged information, the critic predicts the expected success rate of the student's unguided continuation. Thus, a token highly rated by the privileged teacher does not guarantee that the student can successfully navigate the remaining complex steps independently. We posit that a reference solution helps the critic accurately align incomplete responses, improving PPO value estimation beyond standard token-level supervision by bridging the gap between local step correctness and global trajectory success.

We observe different signal profiles for the two readers across response
positions: the critic's prediction benefit grows toward the end, while
self-teacher divergence is high near the beginning, falls through the
middle, and rebounds in the final position bin. This pattern
motivates using the same privileged information for both token guidance and
continuation-value estimation.

Therefore, we introduce \emph{Value-Augmented Self-Distillation Policy Optimization} (VASPO) to leverage shared privileged information for both signals (Figure~\ref{fig:teaser}). In this framework, the actor generates responses conditioned solely on the standard problem prompt. Concurrently, a self-teacher and a critic access the fixed privileged information at the student's sampled prefixes. The critic's value estimates and verifier rewards are used to compute GAE advantages, while clipped teacher--student probability ratios scale selected positive advantages prior to the PPO update. This design effectively integrates privileged information into both the token-guidance and value-estimation paths, all while preserving an autonomous student continuation as the critic's true prediction target.

We evaluate VASPO on mathematical reasoning with Qwen3-4B-Instruct-2507 and OLMo3-7B-Instruct-DPO and on
Knights and Knaves with Qwen3 only. Across these settings, VASPO achieves higher
average accuracy than the evaluated baselines. Further controlled
evaluations examine the contribution of privileged value estimation
and the roles of the two signals across response positions. Our contributions are threefold:
\begin{itemize}
    \item We characterize the positional profiles of privileged information:
    critic prediction benefits grow toward the end, while self-teacher
    divergence is high early, lower in the middle, and rebounds at the end.
    \item We propose VASPO, which shares privileged information between a
    self-teacher and a critic to combine token guidance with
    continuation-value estimation. Critic-derived PPO advantages are
    modulated by bounded teacher preferences, while the actor remains
    prompt-only.
    \item We demonstrate improved mathematical and logical reasoning.
    Across six mathematical benchmarks, VASPO improves the mean score over
    PPO by 2.59 percentage points on Qwen3 and 1.81 points on OLMo3. On
    Knights and Knaves with Qwen3, it also achieves the highest average accuracy
    among the evaluated methods.
\end{itemize}

\section{Preliminaries}
\label{sec:preliminaries}

In this section, we review the two training paradigms that form the basis of our formulation. RLVR provides an outcome-grounded policy-gradient objective, while OPSD offers a privileged token-level signal on the same on-policy prefixes. Finally, we examine how the contribution of privileged information to these two signals evolves across different stages of a generated response.

\subsection{Reinforcement Learning with Verifiable Rewards}
\label{sec:rlvr}

Reinforcement learning with verifiable rewards (RLVR) optimizes a language
model against task outcomes that can be checked automatically, commonly using
PPO-style objectives or group-relative variants such as GRPO
\citep{schulman2017proximalpolicyoptimizationalgorithmsppo,grpo}.  Given a
prompt $x\sim\mathcal{D}_X$, where $\mathcal{D}_X$ is the training prompt
distribution, a policy $\pi_\theta$ generates an autoregressive response
$y=(y_1,\ldots,y_T)$:
\begin{equation}
    y\sim\pi_\theta(\cdot\mid x),
    \qquad h_t=(x,y_{<t}).
    \label{eq:rlvr-rollout}
\end{equation}
A task verifier assigns a scalar terminal reward $\mathcal{R}(x,y)$.  We write
$r_T=\mathcal{R}(x,y)$ and $r_t=0$ for $t<T$, which gives the discounted
return $G_t=\sum_{k=t}^{T}\gamma^{k-t}r_k$ for $\gamma\in(0,1]$.
We optimize the expected terminal reward
\begin{equation}
    J(\theta)
    =\mathbb{E}_{x\sim\mathcal{D}_X,\,y\sim\pi_\theta(\cdot\mid x)}
    \bigl[\mathcal{R}(x,y)\bigr].
    \label{eq:rlvr-objective}
\end{equation}

PPO estimates token-level credit using a value function $V_\phi$ and
generalized advantage estimation (GAE).  With $V_\phi(h_{T+1})=0$, the
temporal-difference residual and advantage are
\begin{equation}
    \begin{aligned}
        \delta_t&=r_t+\gamma V_\phi(h_{t+1})-V_\phi(h_t),\\
        A^{\mathrm{PPO}}_t
        &=\sum_{\ell=0}^{T-t}
            (\gamma\lambda_{\mathrm{GAE}})^\ell\delta_{t+\ell},
    \end{aligned}
    \label{eq:rlvr-gae}
\end{equation}
where $\lambda_{\mathrm{GAE}}\in[0,1]$ controls the bias--variance trade-off.
Let $\pi_{\theta_{\mathrm{old}}}$ be the behavior policy used to collect $y$
and define
$\rho_t(\theta)=\pi_\theta(y_t\mid h_t)/
\pi_{\theta_{\mathrm{old}}}(y_t\mid h_t)$. Write
$\rho_t^{\mathrm{clip}}(\theta)$ for this ratio clipped to
$[1-\epsilon_{\mathrm{clip}},1+\epsilon_{\mathrm{clip}}]$.
The clipped actor loss is
\begin{equation}
    \begin{aligned}
        &\mathcal{L}_{\mathrm{RLVR}}(\theta)\\
        &=-\mathbb{E}_{x,y,t}\!\left[
            \min\!\left(
                \rho_t(\theta)A^{\mathrm{PPO}}_t,
                \rho_t^{\mathrm{clip}}(\theta)A^{\mathrm{PPO}}_t
            \right)
        \right].
    \end{aligned}
    \label{eq:rlvr-loss}
\end{equation}
The expectation averages over training prompts, collected responses, and
valid token positions.  Here the verifier determines the realized outcome,
while the value function provides a prefix-dependent baseline for assigning
token-level credit.

\subsection{On-Policy Self-Distillation}
\label{sec:opsd}

On-policy self-distillation (OPSD) uses privileged information to obtain
token-level supervision without introducing a separate, larger teacher model
\citep{zhao2026opsd}.  Specifically, let $\mathcal{D}$ be a training
distribution over pairs $(x,z)$, whose prompt marginal is $\mathcal{D}_X$ and
where $z$ denotes privileged information available during training.  The
policy samples $y$ as in \eqref{eq:rlvr-rollout}.  At each resulting prefix
$h_t$, OPSD evaluates a student distribution and a detached privileged
self-teacher distribution:
\begin{equation}
    p_{\theta,t}=\pi_\theta(\cdot\mid h_t),
    \qquad
    q^z_{\bar\theta,t}
    =\operatorname{sg}\!\left[
        \pi_{\bar\theta}(\cdot\mid h_t,z)
    \right],
    \label{eq:opsd-distributions}
\end{equation}
where $\operatorname{sg}$ denotes stop-gradient and $\bar\theta$ denotes the
parameters of the self-teacher copy.  Standard OPSD trains the policy by
minimizing a token-wise divergence along its own trajectories:
\begin{equation}
    \begin{aligned}
        &\mathcal{L}_{\mathrm{OPSD}}(\theta)\\
        &=\mathbb{E}_{\substack{(x,z)\sim\mathcal{D}\\
                               y\sim\pi_\theta(\cdot\mid x)}}
        \!\left[
            \frac{1}{T}\sum_{t=1}^{T}
            D\!\left(q^z_{\bar\theta,t}\,\|\,p_{\theta,t}\right)
        \right],
    \end{aligned}
    \label{eq:opsd-objective}
\end{equation}
where $D$ is a distributional divergence and gradients propagate only through
$p_{\theta,t}$.  RLVR and OPSD thus evaluate the same
prefixes using complementary signals: a verifier-grounded return and a
privileged next-token distribution, respectively.

\subsection{The Positional Divide of Supervision}
\label{sec:position-motivation}

\begin{figure}[H]
    \centering
    \includegraphics[width=\linewidth]{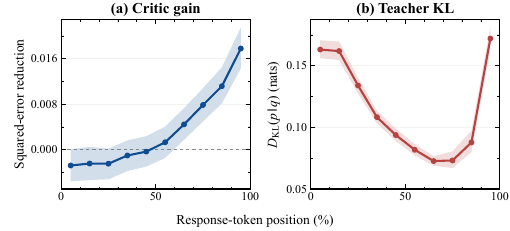}
    \caption{\textbf{Privileged signals by relative response position.}
    All responses:
    (a) solution-induced reduction in critic squared
    error (positive is better); (b) $D_{\mathrm{KL}}(p\|q)$ between the clean
    actor and privileged self-teacher. Shading shows 95\% confidence
    intervals.}
    \label{fig:position-signals}
\end{figure}

We compare ordinary and solution-conditioned critics on the same fixed
responses, alongside the actor acting as its own solution-conditioned
teacher. Appendix~\ref{app:frozen-response-position} provides the
experimental details.

Figure~\ref{fig:position-signals} shows different positional profiles:
critic prediction benefits grow toward the end, while teacher divergence
is high early, falls through the middle, and rebounds at the end. These profiles motivate using privileged information in both readers:
the critic forms outcome-grounded advantages, and the self-teacher
modulates token updates.
\section{Methodology}
\label{sec:method}

\begin{figure*}[t]
    \centering
    \includegraphics[width=\textwidth]{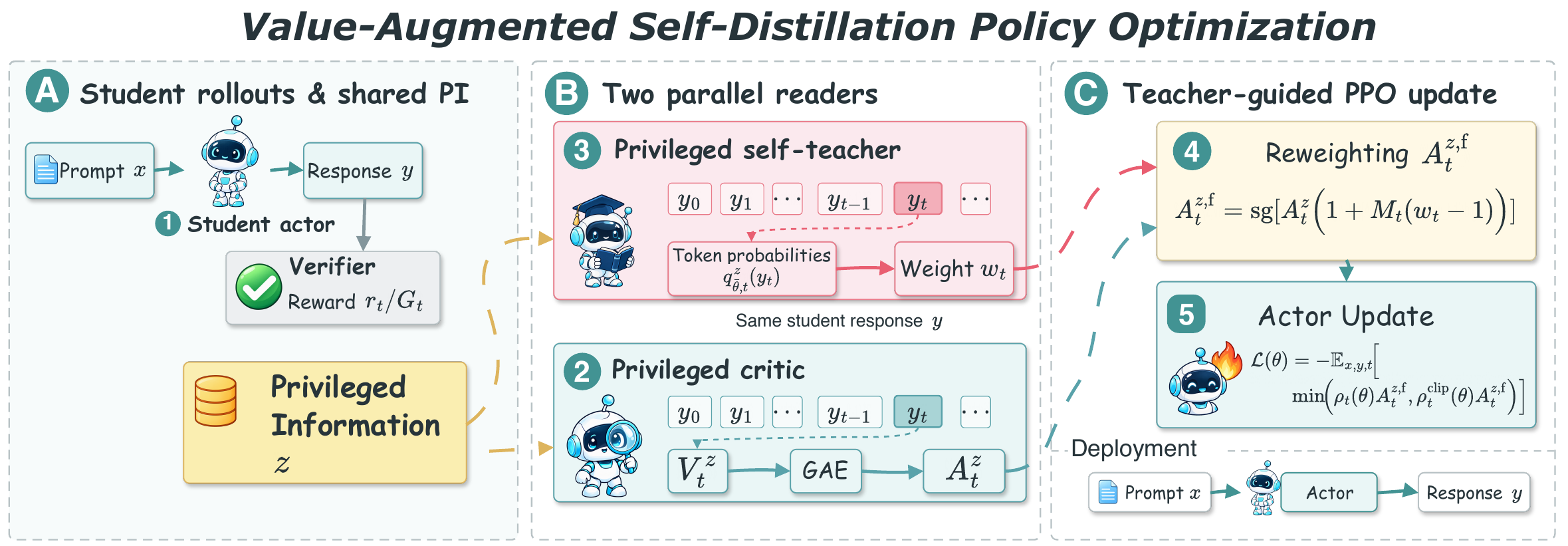}
    \caption{\textbf{Overview of VASPO training and deployment.}
    Shared privileged information (PI) supplies critic values for GAE and
    teacher weights for advantage modulation. The actor remains prompt-only.}
    \label{fig:port-overview}
\end{figure*}

\subsection{Overview of VASPO}
\label{sec:method-overview}

VASPO leverages shared privileged information for both value estimation and self-distillation, while ensuring the student model operates solely on the standard prompt during training and inference. As illustrated in Figure~\ref{fig:port-overview}, the framework retains the clipped actor update of PPO but introduces two privileged readers with distinct roles. The privileged critic estimates the expected return of an autonomous student continuation, supplying the values required to compute token-level GAE advantages. Concurrently, the self-teacher evaluates the student's sampled tokens using the same privileged context; its preferences are transformed into bounded coefficients that scale selected positive advantages. This privileged information is established prior to rollout and accessed by both readers at identical student-generated prefixes. The remainder of this section is organized as follows: Section~\ref{sec:privilege-bank} formalizes the shared privileged information and student rollouts, Section~\ref{sec:privileged-critic} details the privileged value estimation, and Section~\ref{sec:privileged-teacher} presents the teacher-guided reweighting alongside the actor update. Algorithm~\ref{alg:shared-privilege} summarizes the complete VASPO training procedure.

\subsection{Shared Privileged Information}
\label{sec:privilege-bank}

The privileged information is shared by the teacher and critic, while the
student learns from its own responses. For a prompt $x$, we select privileged
information $z$ before rollout and keep it fixed. The student samples $y$ from
$\pi_{\theta_{\mathrm{old}}}(\cdot\mid x)$. At each prefix
$h_t=(x,y_{<t})$, both readers receive the same $z$ and prefix; the student
receives no privileged information during either rollout or deployment.

\subsection{Privileged Value Estimation}
\label{sec:privileged-critic}

\paragraph{Continuation-value prediction.}
The critic uses the privileged information to estimate the return of a
continuation generated by the student without privileged input. A worked
solution may help it relate an unfinished response to the steps the task
requires. Its input is the prefix $h_t$ and privileged information $z$, and
its prediction target is
\begin{equation}
    V^z_\phi(h_t,z)
    \approx
    \mathbb{E}_{\pi_{\theta_{\mathrm{old}}}}[G_t\mid h_t,z].
    \label{eq:privileged-value}
\end{equation}
The expectation is over continuations from the student that collected the
rollout. For binary success rewards without discounting, this is the
probability that the student will complete correctly. The critic does not
observe the current token or the continuation it is predicting. The privileged
information supplies task information; the student's observed return supplies
the training target.

\paragraph{Critic pre-update.}
We update the critic on each newly collected rollout batch before computing
the advantages used by the actor. With the student held fixed at
$\pi_{\theta_{\mathrm{old}}}$, the verifier supplies the terminal return
$G_i=\mathcal{R}(x_i,y_i)$ for response $i$. We use $\gamma=1$, so every
valid token in this response has the same return target $G_i$. This update
fits the value estimator to the latest student trajectories before its
predictions determine token credit. The privileged information enters the
critic input, while supervision remains the student's observed return.

\paragraph{PPO clipped value loss.}
Let $v^z_{i,t}=V^z_\phi(h_{i,t},z_i)$ be the current value prediction and
$v^{z,\mathrm{old}}_{i,t}$ the prediction stored before the critic
pre-update. With value clip range $\epsilon_v$, we define
\begin{equation}
    \bar v^z_{i,t}
    =\operatorname{clip}\!\left(
        v^z_{i,t},
        v^{z,\mathrm{old}}_{i,t}-\epsilon_v,
        v^{z,\mathrm{old}}_{i,t}+\epsilon_v
      \right).
    \label{eq:clipped-privileged-value}
\end{equation}
The critic minimizes the PPO clipped value loss
\begin{equation}
    \begin{aligned}
        \mathcal{L}_V(\phi)
        =\frac{1}{2}\operatorname{Agg}_{m}\!\Bigl[
          \max\!\bigl(
            &(v^z_{i,t}-G_i)^2,\\
            &(\bar v^z_{i,t}-G_i)^2
          \bigr)
        \Bigr],
    \end{aligned}
    \label{eq:privileged-value-loss}
\end{equation}
where $\operatorname{Agg}_{m}$ aggregates over valid response tokens under
mask $m$. Both the return targets and the stored predictions are detached
and fixed throughout this critic update phase. We perform one such phase
per rollout batch and denote the resulting parameters by $\phi^+$.

\paragraph{Advantages from refreshed values.}
After the pre-update, we evaluate the same prefixes again with
$V^z_{\phi^+}$ and recompute $A^z_t$ using the GAE definition in
\eqref{eq:rlvr-gae}, with the refreshed privileged critic as the value
function.
With $V^z_{\phi^+}(h_{T+1},z)=0$, terminal-only rewards, and
$\gamma=\lambda_{\mathrm{GAE}}=1$, the GAE sum telescopes, yielding
\begin{equation}
    A^z_{i,t}=G_i-v^{z,+}_{i,t},
    \qquad v^{z,+}_{i,t}=V^z_{\phi^+}(h_{i,t},z_i).
    \label{eq:privileged-advantage}
\end{equation}
Although $G_i$ is shared across the response, the refreshed baseline varies
with the prefix, retaining PPO's token-level credit structure. We detach
these advantages and hold them fixed during the subsequent actor update.
There is no second critic update on this batch. This
\emph{privileged-critic advantage} combines the verifier outcome with a
prefix-dependent value informed by privileged information, before any
teacher reweighting.

\subsection{Teacher-Guided PPO Update}
\label{sec:privileged-teacher}
\label{sec:advantage-interface}

\paragraph{From teacher scores to token weights.}
The self-teacher evaluates the same tokens sampled by the student, using
the distribution conditioned on privileged information $q^z_{\bar\theta,t}$ from
\eqref{eq:opsd-distributions}. Let
$p_{\mathrm{old},t}=\pi_{\theta_{\mathrm{old}}}(\cdot\mid h_t)$ be the
student's behavior distribution. We compute the detached student--teacher
gap on the same sampled token $y_t$:
\begin{equation}
    d_t=\operatorname{sg}\!\left[
        \log q^z_{\bar\theta,t}(y_t)-\log p_{\mathrm{old},t}(y_t)
    \right].
    \label{eq:teacher-gap}
\end{equation}
We clip the corresponding probability ratio to obtain a weight:
\begin{equation}
    w_t=\operatorname{clip}\!\left(
        e^{d_t},1-\epsilon_w,1+\epsilon_w\right).
    \label{eq:teacher-weight}
\end{equation}
Here $0<\epsilon_w<1$ bounds the reweighting. A token assigned more
probability by the teacher receives a weight above
one; a token assigned less receives a weight below one. Clipping bounds
how much this preference can change the critic-derived credit.

\paragraph{Token-level mask.}
Teacher modulation is restricted to tokens with positive GAE advantages
and a sufficiently large teacher--student gap. The advantage gate selects
tokens that the outcome-grounded update already encourages; the gap gate
leaves near-agreement tokens at their original weight. For each valid
response token, we define
\begin{equation}
    M_t=\mathbf{1}[A^z_t>0]\,
        \mathbf{1}\!\left[|\tanh(d_t/\tau)|>\delta\right].
    \label{eq:teacher-token-mask}
\end{equation}
The sign test uses the raw, unwhitened $A^z_t$ recomputed after the critic
pre-update. It is applied at each token position, without directly checking
whether the complete response is correct. Prompt and padding positions are
excluded by the valid-response-token mask. Thus, $M_t$ controls where teacher
guidance changes the actor advantage; it does not select the examples used
to train the critic.

\paragraph{Reweighting the advantage.}
We denote the fused advantage by $A^{z,\mathrm{f}}_t$, where
$\mathrm{f}$ indicates teacher fusion, and combine the refreshed GAE
advantage, token mask, and teacher weight as
\begin{equation}
    A^{z,\mathrm{f}}_t=\operatorname{sg}\!\left[
        A^z_t\bigl(1+M_t(w_t-1)\bigr)\right].
    \label{eq:direction-preserving-fusion}
\end{equation}
For selected tokens this gives $A^{z,\mathrm{f}}_t=A^z_tw_t$; for all other tokens it gives
$A^{z,\mathrm{f}}_t=A^z_t$. The teacher therefore strengthens or attenuates selected
positive updates, while negative advantages and unselected tokens retain
their original credit. The positive weight preserves the sign of $A^z_t$.
We use the detached $A^{z,\mathrm{f}}_t$ in the clipped PPO loss
\eqref{eq:rlvr-loss} over all valid response tokens.
Algorithm~\ref{alg:shared-privilege} summarizes the training procedure.

\begin{algorithm}[tb]
\small
\caption{VASPO Training Pipeline}
\label{alg:shared-privilege}
\begin{algorithmic}[1]
\Require Data $\mathcal{D}$, student $\pi_\theta$, EMA teacher
$\pi_{\bar\theta}$, critic $V^z_\phi$, verifier $\mathcal{R}$
\For{each training iteration}
    \State Set $\theta_{\mathrm{old}}\gets\theta$.
    \State Sample $\{(x_i,z_i)\}_{i=1}^{B}\sim\mathcal{D}$; generate
    $y_i\sim\pi_{\theta_{\mathrm{old}}}(\cdot\mid x_i)$ and obtain
    returns $G_i=\mathcal{R}(x_i,y_i)$.
    \State Store $v^{z,\mathrm{old}}_{i,t}$; pre-update the critic to $\phi^+$ using
    Eq.~\ref{eq:privileged-value-loss}.
    \State Re-score $V^z_{\phi^+}$; recompute $A^z_{i,t}$ via
    Eq.~\ref{eq:privileged-advantage}.
    \State Compute teacher weights $w_{i,t}$ and token masks $M_{i,t}$
    using Eqs.~\ref{eq:teacher-gap}--\ref{eq:teacher-token-mask}.
    \State Form $A^{z,\mathrm{f}}_{i,t}$ using Eq.~\ref{eq:direction-preserving-fusion}.
    \State Update $\pi_\theta$ with fixed $A^{z,\mathrm{f}}_{i,t}$ using
    Eq.~\ref{eq:rlvr-loss}.
    \State Set $\phi\gets\phi^+$ for the next iteration.
\EndFor
\end{algorithmic}
\end{algorithm}

\section{Experiments}
\label{sec:experiments}

\begin{figure*}[t]
    \centering
    \includegraphics[width=\textwidth]{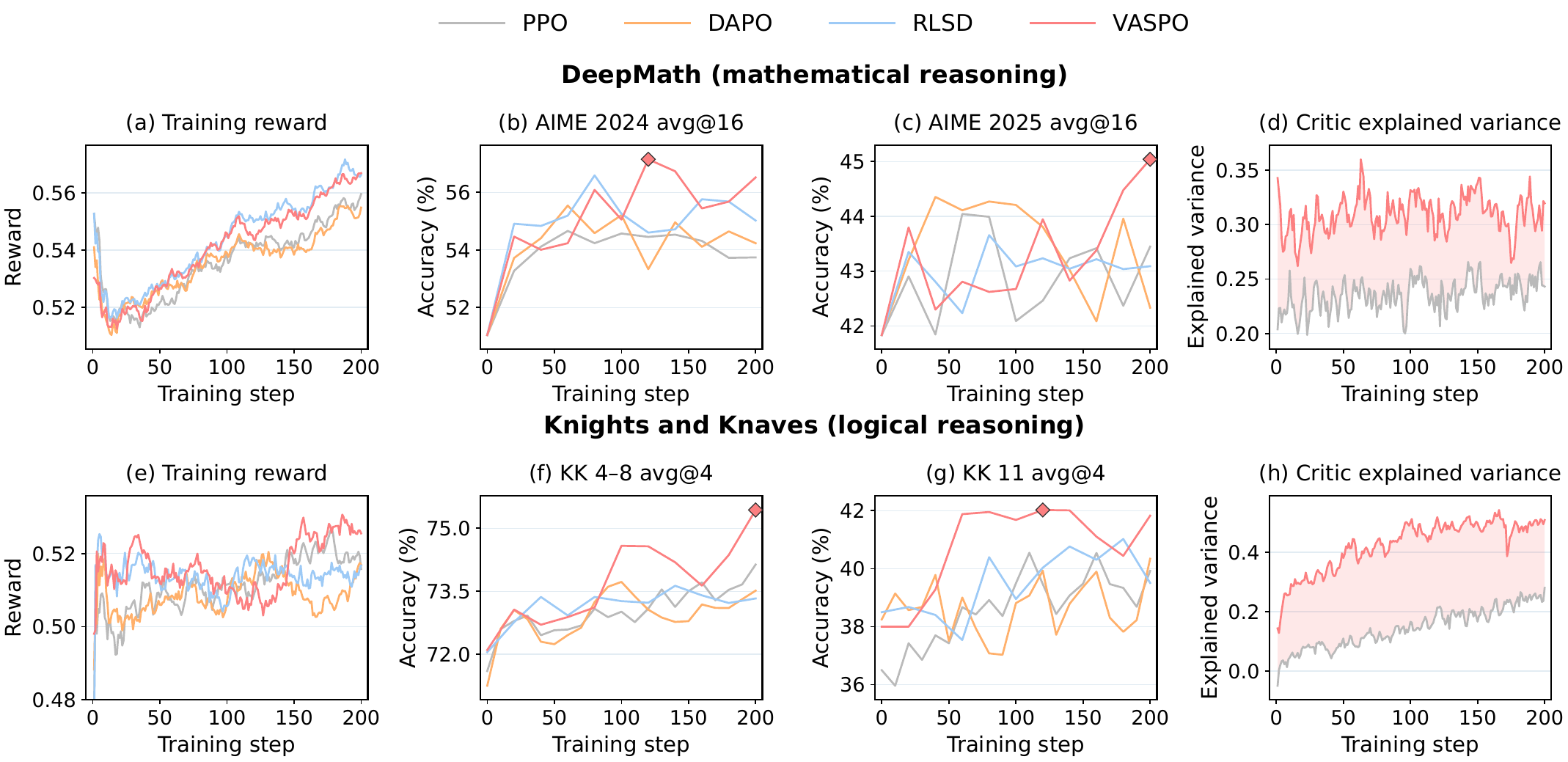}
    \caption{\textbf{Training dynamics of Qwen3-4B-Instruct-2507.}
    Top: DeepMath training reward, AIME 2024 and AIME 2025 avg@16 accuracy,
    and critic explained variance. Bottom: KK training reward, 4--8-role
    and 11-role avg@4 accuracy, and critic explained variance.}
    \label{fig:training-dynamics}
\end{figure*}

\subsection{Experimental Setup}
\label{sec:experimental-setup}

\paragraph{Models and Datasets.}
We conduct experiments on Qwen3-4B-Instruct-2507 \citep{yang2025qwen3}
and OLMo3-7B-Instruct-DPO \citep{olmo2025olmo3} for mathematical reasoning,
and on Qwen3-4B-Instruct-2507 for logical reasoning. We train on the full
DeepMath-103K dataset \citep{he2025deepmath} for mathematics and Knights
and Knaves (KK) \citep{xie2025logic} puzzles with 4--8 roles for logic.
Privileged information consists of the solutions and answers provided in
the datasets.

\paragraph{Baselines and Evaluation.}
We compare VASPO with Vanilla (the initial checkpoint), DAPO, PPO
\citep{schulman2017proximalpolicyoptimizationalgorithmsppo},
OPSD \citep{zhao2026opsd}, and RLSD
\citep{yang2026selfdistilledrlvr}. For mathematics, we report avg@32 on
AIME24, AIME25, and AIME26 \citep{maaAime} and HMMT25,
and avg@8 on Minerva Math \citep{lewkowycz2022minerva}
and OlympiadBench \citep{he2024olympiadbench}. For KK, we report avg@16
on held-out 4--8-role puzzles
and out-of-distribution tests with 9, 10, and 11 roles. Avg@$K$ denotes
mean correctness over $K$ sampled responses per problem.
We assess critic quality using explained variance (EV), which measures
how much return variation is explained by its value predictions:
\begin{equation}
    \mathrm{EV}
    =1-\frac{\operatorname{Var}_{\mathcal{M}}(G_t-V_t)}
                  {\operatorname{Var}_{\mathcal{M}}(G_t)+\varepsilon_{\mathrm{EV}}},
    \label{eq:critic-ev}
\end{equation}
where $\mathcal{M}$ contains valid response-token positions, $V_t$ is the
raw critic prediction, $G_t$ is the clean student's verifier-return target,
and $\varepsilon_{\mathrm{EV}}>0$ ensures numerical stability.
Higher EV indicates lower residual variance relative to return variance.

\paragraph{Implementation Details.}
We implement VASPO in \texttt{verl} \citep{sheng2024hybridflow}, using
full-parameter training with AdamW and actor/critic learning rates of
$10^{-6}$/$5\times10^{-6}$. The global batch size is 128 and the mini-batch
size is 32, with eight responses sampled per prompt. We use $\gamma=1$,
$\lambda_{\mathrm{GAE}}=1$, PPO clipping $\epsilon_{\mathrm{clip}}=0.2$,
reweighting clip $\epsilon_w=0.2$, and selection parameters
$\tau=\delta=0.02$. Training and evaluation use temperatures of $1.0$
and $0.6$, respectively, with top-$p$ $0.95$ and top-$k$ $20$.
The maximum response length is $8{,}192$ tokens for both training and
evaluation. Critic pretraining, evaluation, and training settings are
provided in Appendices~\ref{app:critic-pretraining},
\ref{app:evaluation-details}, and~\ref{app:training-details}, respectively.
For Qwen3-4B-Instruct-2507, student rollout is the most time-consuming
component, averaging
$191.97$\,s per step over 200 training steps
(Appendix~\ref{app:training-time}).

\subsection{Main Results}
\label{sec:main-results}

\begin{table*}[!t]
    \centering
    \begingroup
    \small
    \setlength{\tabcolsep}{4pt}
    \resizebox{\textwidth}{!}{    \begin{tabular}{lccccccc}
        \toprule
        \multirow{2}{*}{\textbf{Method}} & \textbf{AIME24} & \textbf{AIME25} & \textbf{AIME26} & \textbf{HMMT25} & \textbf{Minerva} & \textbf{Olympiad} & \multirow{2}{*}{\textbf{Avg.}} \\
        \cmidrule(lr){2-5}\cmidrule(lr){6-7}
        & \multicolumn{4}{c}{\textbf{avg@32}} & \multicolumn{2}{c}{\textbf{avg@8}} & \\
        \midrule
        \textbf{Qwen3-4B-Instruct-2507} & & & & & & & \\
        \quad Vanilla & 51.04 & 40.83 & 48.85 & 25.83 & 60.43 & 71.55 & 49.76 \\
        \quad DAPO & 53.85 & 42.08 & \underline{49.90} & \underline{28.33} & 62.59 & \underline{72.81} & \underline{51.59} \\
        \quad PPO & 51.67 & \underline{42.71} & 48.65 & 26.35 & \underline{63.14} & 70.96 & 50.58 \\
        \quad OPSD & 29.58 & 26.46 & 31.77 & 12.29 & 54.14 & 54.56 & 34.80 \\
        \quad RLSD & \underline{54.17} & 42.60 & 49.17 & 27.08 & 60.11 & 72.14 & 50.88 \\
        \rowcolor{blue!5}
        \quad \textbf{VASPO} & \textbf{56.87} & \textbf{43.96} & \textbf{51.15} & \textbf{29.17} & \textbf{64.75} & \textbf{73.12} & \textbf{53.17} \\
        \midrule
        \textbf{OLMo3-7B-Instruct-DPO} & & & & & & & \\
        \quad Vanilla & 24.17 & 20.62 & 21.15 & 11.35 & 45.36 & 56.21 & 29.81 \\
        \quad DAPO & 27.92 & 23.13 & \underline{23.23} & 13.65 & 47.01 & \underline{58.64} & 32.26 \\
        \quad PPO & 25.10 & \underline{24.38} & 22.40 & 12.71 & 47.06 & 58.01 & 31.61 \\
        \quad OPSD & 18.33 & 16.15 & 15.31 & 10.52 & 35.11 & 46.61 & 23.67 \\
        \quad RLSD & \underline{28.44} & 23.65 & 22.71 & \underline{14.69} & \underline{47.08} & 58.25 & \underline{32.47} \\
        \rowcolor{blue!5}
        \quad \textbf{VASPO} & \textbf{29.58} & \textbf{24.58} & \textbf{24.90} & \textbf{14.90} & \textbf{47.75} & \textbf{58.83} & \textbf{33.42} \\
        \bottomrule
    \end{tabular}    }
    \endgroup
    \caption{Mathematical reasoning accuracy (\%).
    Bold/underlined: best/second-best available scores per backbone.}
    \label{tab:main-results}
\end{table*}

\begin{table}[!t]
    \centering
    \begingroup
    \small
    \setlength{\tabcolsep}{3pt}
    \resizebox{\columnwidth}{!}{    \begin{tabular}{lccccc}
        \toprule
        & \multicolumn{1}{c}{ID} & \multicolumn{3}{c}{OOD} & \\
        \cmidrule(lr){2-2}\cmidrule(lr){3-5}
        \textbf{Method} & \textbf{4--8 Roles} & \textbf{9 Roles} & \textbf{10 Roles} & \textbf{11 Roles} & \textbf{Avg.} \\
        \midrule
        Vanilla & 72.99 & 67.94 & 54.44 & 37.81 & 58.30 \\
        DAPO & 73.71 & 68.81 & 57.00 & 40.00 & 59.88 \\
        PPO & \underline{74.33} & 65.59 & \underline{57.69} & 39.14 & 59.19 \\
        RLSD & 73.54 & \textbf{70.19} & 57.50 & \underline{40.12} & \underline{60.34} \\
        \rowcolor{blue!5}
        \textbf{VASPO} & \textbf{75.31} & \underline{69.25} & \textbf{58.06} & \textbf{40.62} & \textbf{60.81} \\
        \bottomrule
    \end{tabular}    }
    \endgroup
    \caption{Knights and Knaves accuracy (avg@16, \%) on
    Qwen3-4B-Instruct-2507. Avg.: mean across the four role groups.
    Bold/underlined: best/second-best.}
    \label{tab:logic-results}
\end{table}

\paragraph{Mathematical reasoning.}
VASPO surpasses all baselines across the mathematical benchmarks on both Qwen3 and OLMo3 (Table~\ref{tab:main-results}). The performance gains hold across various tasks and architectures. DAPO and RLSD serve as the strongest average baselines on Qwen3 and OLMo3, respectively, while OPSD universally declines. This indicates that raw access to privileged information is insufficient for better reasoning.

\paragraph{Logical reasoning and transfer.}
Furthermore, VASPO sets the highest average performance on the KK dataset, improving both in-domain reasoning and out-of-distribution transfer to larger puzzles (Table~\ref{tab:logic-results}). It leads the training-range split and the largest out-of-distribution splits, though RLSD edges it out on 9-role puzzles. Overall, the margin of improvement is smaller than in mathematics, and all methods suffer performance drops as puzzle size scales.

\paragraph{Training dynamics.}
VASPO's higher held-out accuracy is accompanied by better critic predictions
despite similar training rewards (Figure~\ref{fig:training-dynamics}). It
finishes above PPO, DAPO, and RLSD on the AIME benchmarks and KK splits shown,
while its critic attains higher explained variance than PPO. The separation
in held-out performance is therefore more pronounced than in training reward.
This pattern is consistent with the hypothesis that privileged value
estimation supports policy learning; the component ablations and value
diagnostics below examine this explanation more directly.

\subsection{Ablation Study}
\label{sec:ablations}

\begin{table*}[t]
    \centering
    \small
    \setlength{\tabcolsep}{3pt}
    \resizebox{\textwidth}{!}{    \begin{tabular}{lcccccccccc}
        \toprule
        & \multicolumn{5}{c}{Mathematical Reasoning}
        & \multicolumn{5}{c}{Logical Reasoning: Knights and Knaves} \\
        \cmidrule(lr){2-6}\cmidrule(lr){7-11}
        Method & AIME2024 & AIME2025 & AIME2026 & Olympiad & Avg.
        & 4--8 Roles & 9 Roles & 10 Roles & 11 Roles & Avg. \\
        \midrule
        \rowcolor{blue!5}
        \textbf{VASPO (full)} & \textbf{56.87} & \textbf{43.96} & \textbf{51.15} & \textbf{73.12} & \textbf{56.28}
        & \textbf{75.31} & \textbf{69.25} & \textbf{58.06} & \textbf{40.62} & \textbf{60.81} \\
        $-$ privileged critic & 29.58 & 26.46 & 31.77 & 54.56 & 35.59
        & 47.20 & 25.19 & 15.06 & 5.75 & 23.30 \\
        $-$ self-teacher modulation & 55.73 & 43.44 & 50.73 & 73.07 & 55.74
        & 73.95 & 69.25 & 56.13 & 39.25 & 59.65 \\
        $-$ both reader components & 51.67 & 42.71 & 48.65 & 70.96 & 53.50
        & 74.33 & 65.59 & 57.69 & 39.14 & 59.19 \\
        $-$ token-level mask & 55.41 & 41.56 & 50.42 & 72.40 & 54.95
        & 74.84 & 67.19 & 56.94 & 39.44 & 59.60 \\
        \midrule
        $-$ solution derivation (answer only) & 54.90 & 41.88 & 49.48 & 72.11 & 54.59
        & 73.46 & 66.75 & 56.69 & 38.75 & 58.91 \\
        $-$ critic pre-update & 55.52 & 42.71 & 49.17 & 72.59 & 55.00
        & 74.35 & 67.00 & 55.31 & 38.69 & 58.84 \\
        $+$ advantage whiten & 54.17 & 42.50 & 48.44 & 72.53 & 54.41
        & 73.84 & 66.19 & 54.63 & 38.31 & 58.24 \\
        \bottomrule
    \end{tabular}    }
    \caption{Ablation of critic, self-teacher and privileged-information content
    in VASPO (Qwen3-4B-Instruct-2507; accuracy, \%). Rows denote changes
    relative to the full method.
    Mathematics uses avg@32 on AIME2024/2025/2026
    and avg@8 on OlympiadBench; KK uses avg@16, with 4--8 roles and
    9/10/11 roles.}
    \label{d}
    \label{tab:reader-effects}
    \label{tab:privileged-information}
\end{table*}

\paragraph{Contributions of the teacher and critic.}VASPO yields the highest average accuracy among all variants in both mathematics and KK (Table~\ref{tab:reader-effects}). The privileged critic alone improves performance over PPO, and adding self-teacher modulation provides further gains. Across all benchmarks, the full method matches or outperforms the critic-only variant. The variant without the privileged critic removes the entire critic component , leaving OPSD as the training objective. Its lower accuracy supports retaining the critic-based reinforcement-learning path alongside self-teacher guidance.

\paragraph{Ablation on token-level mask.}
Selective teacher modulation consistently improves performance (Table~\ref{tab:reader-effects}). Removing the token-level mask in \eqref{eq:teacher-token-mask} degrades accuracy across every benchmark and role group in both domains.

\begin{figure}[t]
    \centering
    \includegraphics[width=\columnwidth]{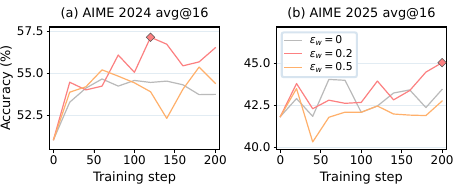}
    \caption{Fusion strength on Qwen3-4B-Instuct-2507.
    AIME 2024 and AIME 2025 avg@16 accuracy over DeepMath training steps.
    The probability-ratio clip width
    $\epsilon_w=0$ disables teacher modulation.}
    \label{fig:fusion-lambda}
\end{figure}

\paragraph{Ablation on fusion strength.}Stronger teacher modulation does not guarantee better performance (Figure~\ref{fig:fusion-lambda}). Modulated by the probability-ratio clip width $\epsilon_w$ in \eqref{eq:teacher-weight}, the default $\epsilon_w=0.2$ achieves higher final AIME accuracy than either disabling it (zero) or widening it to $0.5$.

\paragraph{Ablation on privileged-information content.}

\label{sec:evidence-cost}

Replacing the solution-plus-answer input with an answer-only variant lowers accuracy across all evaluated benchmarks and role groups (Table~\ref{tab:privileged-information}). Both readers receive the modified input, while the actor remains conditioned on the standard prompt.

\paragraph{Critic pre-update.}
Removing the critic pre-update incurs a consistent accuracy drop across all mathematical benchmarks and KK role groups (Table~\ref{tab:reader-effects}). Because self-teacher advantage modulation can amplify actor updates, it risks exacerbating the distribution shift between the evolving policy and a critic fitted to outdated rollouts. Updating the critic on the latest actor trajectories prior to computing advantages enables it to align with the current policy before the subsequent actor update. This timely adaptation enhances value-prediction accuracy and yields more reliable token-level credit, effectively explaining the observed performance gains.

\paragraph{Advantage whitening.}
Applying global whitening after teacher modulation degrades performance across all mathematical benchmarks and KK role groups (Table~\ref{tab:reader-effects}). Although strictly positive teacher weights preserve the signs of the raw advantages, the subsequent subtraction of a batch-wide mean can inadvertently invert the signs of some fused advantages. Consequently, tokens initially encouraged by the outcome-grounded signal might receive penalizing updates, or vice versa, severely distorting the intended optimization directions. This phenomenon accounts for the observed degradation and strongly advocates for utilizing the fused advantages without any subsequent global whitening.

\subsection{Analysis}
\label{sec:analysis}

\paragraph{Does privileged information improve value prediction?}
\label{sec:value-diagnostic}
Conditioning on reference solutions improves value prediction point estimates but does not aid calibration (Figure~\ref{fig:value-diagnostics}). Evaluated on fixed continuations from a frozen Qwen3 actor, the privileged solution critic yields the highest explained variance and lowest Brier error across the three checkpoints, whereas the baseline critic maintains the lowest ECE. The point-estimate rankings alone do not establish a strict overall advantage.

\paragraph{Where do the two signals help?}
The critic's benefit varies significantly by prefix position and source-response correctness. Aggregate diagnostics indicate that solution conditioning degrades accuracy at early prefixes but improves it at late prefixes, with mixed results in the middle. Paired positional contrasts confirm this trend, highlighting that late-stage gains are largest for incorrect source responses. While length controls validate this pattern for medium and long responses, the late gain remains uncertain in the long-response group (Appendix~\ref{app:position-analysis}). Conversely, the self-teacher follows a different trajectory: its divergence from the baseline actor is high initially, falls through the middle, and rebounds at the final position (Figure~\ref{fig:position-signals}). These complementary signal profiles motivate employing the two readers for distinct roles in the update.

\section{Conclusion}
\label{sec:conclusion}

We presented VASPO, a PPO framework leveraging shared reference solutions for continuation-value estimation and token guidance. By combining privileged critic credit and bounded self-teacher modulation, VASPO trains actors using solely standard prompts. Empirically, it achieves top average accuracy across Qwen3 and OLMo3 mathematical and logical reasoning benchmarks. Analyses show privileged conditioning trades higher early-prefix error and worse calibration for improved late-prefix value prediction. Lastly, ablations confirm that full solution derivations and selective, moderate teacher modulation are crucial.

\section*{Limitations}
\label{sec:limitations}

Extending VASPO to interactive tasks requires careful consideration of when newly observed information becomes available to the value estimator. Specifically, under the current protocol, environmental feedback obtained at the completion of a response cannot be retroactively used as privileged context for earlier prefixes of that same trajectory.

Furthermore, the framework's efficacy is constrained by evidence quality and computational overhead. Privileged reference solutions, even when yielding a verified final answer, may still harbor flawed intermediate reasoning steps that could introduce noisy supervision. Additionally, conditioning the readers on full solutions expands the context window for both the teacher and the critic, thereby increasing training costs.

Finally, VASPO is currently designed for single-turn generation. Its application to multi-turn dialogue---where conversational history and available privileged evidence dynamically evolve across successive turns---remains an open direction for future work.

\appendix
\section{LLM Usage}

We used LLMs only to polish grammar and improve the clarity of the manuscript. All research ideas, experiments, and analyses were conducted by the authors.

\section{Related Work}
\label{app:related-work}

\subsection{RLVR and Value-Based Credit Assignment}

Verifiable outcomes provide a scalable training signal for reasoning, while
the policy optimizer determines how that signal is assigned to sampled tokens.
DeepSeekMath introduced GRPO, which compares rewards within a group of
responses to estimate advantages without a learned critic \citep{grpo};
DeepSeek-R1 further demonstrated the potential of outcome-based reinforcement
learning for reasoning \citep{guo2025deepseek}.  In outcome-only GRPO, a
response-level advantage is shared across token positions.  PPO instead uses
a learned value function to construct prefix-dependent advantages
\citep{schulman2017proximalpolicyoptimizationalgorithmsppo}.  Process
supervision provides another route to finer-grained feedback by learning from
intermediate-step correctness labels \citep{lightman2023let}.  Step-potential
advantage estimation similarly derives step-level credit from intermediate
confidence and correctness signals \citep{wu2026steppotential}.  VASPO retains
terminal verification as the source of rewards and value targets, and uses
training-time evidence to inform the two readers that construct token-level
updates.

Recent actor--critic methods study value estimation for
long reasoning trajectories.  VC-PPO pretrains the critic with Monte Carlo
returns and decouples the GAE parameters used for value and policy updates
\citep{yuan2025vcppo}.  VAPO builds on these techniques and adds
length-adaptive advantage estimation to address heterogeneous response
lengths \citep{yue2025vapo}.  GenAC replaces a scalar value head with a generative
critic that reasons about a prefix before predicting its value
\citep{shan2026genac}.  ReDiPPO instead calibrates value estimates with
reference-model guidance and reweights tokens according to policy--reference
discrepancy \citep{zhang2026redippo}.  In OPD, \citet{oh2026klbaseline} derive a closed-form
reverse-KL control-variate baseline to reduce gradient variance without an
additional critic. That baseline belongs to the distillation objective;
VASPO's learned critic predicts verifier returns from clean student
continuations. VASPO studies the critic's input choice: conditioning this
return estimate on the evidence already supplied to a self-teacher.

\subsection{On-Policy Distillation and Self-Teaching}

On-policy distillation supplies teacher feedback on the states that the
student actually visits.  GKD trains on student-generated sequences and
studies several distribution-matching objectives \citep{agarwal2024policy},
while MiniLLM uses a reverse-KL objective optimized through policy gradients
\citep{gu2024minillm}.  More recent work also considers objectives that relax
strict imitation: ExOPD scales the teacher-derived implicit reward within a
KL-constrained formulation \citep{yang2026learning}.  These methods establish
student-generated prefixes as the interface for dense teacher supervision.

Self-distillation obtains a better-informed teacher by changing its context
rather than requiring a separate, larger model.  Retrieval-augmented
generation conditions a model on external evidence in a similar spirit:
Stable-RAG shows that permuting the retrieved clues can by itself induce
hallucinations \citep{zhang2026stable}, while compact clue selection reduces
such evidence to an informative subset for efficient reasoning
\citep{zhang2026less}.  This use of training-only
context connects self-teaching to learning using privileged information and
generalized distillation \citep{vapnik2009lupi,lopezpaz2016generalized}. OPSD
conditions a detached self-teacher on worked solutions and matches its
distributions along clean student rollouts \citep{zhao2026opsd}.  SDPO
constructs self-teaching contexts from rich environment feedback and
successful attempts, allowing the policy to learn from information revealed
during interaction
\citep{hubotter2026reinforcement}.  Studies of privileged OP(S)D identify
failure modes for thinking models, instance-specific privileged information,
and unreliable teacher targets at some student prefixes
\citep{kaur2026rethinking,zhu2026manyfaces,liu2026teachertokens}. These
findings concern the teacher signal and its downstream effect on the actor.
VASPO uses the privileged information for sampled-token assessment and
additionally tests its effect on continuation-value estimation. This second
role constrains evidence timing: the shared bank is fixed before rollout,
whereas feedback from the current completed response can inform a
retrospective teacher without being an admissible input to VASPO's critic for
that same response.

\begin{figure*}[t]
    \centering
    \includegraphics[width=\textwidth]{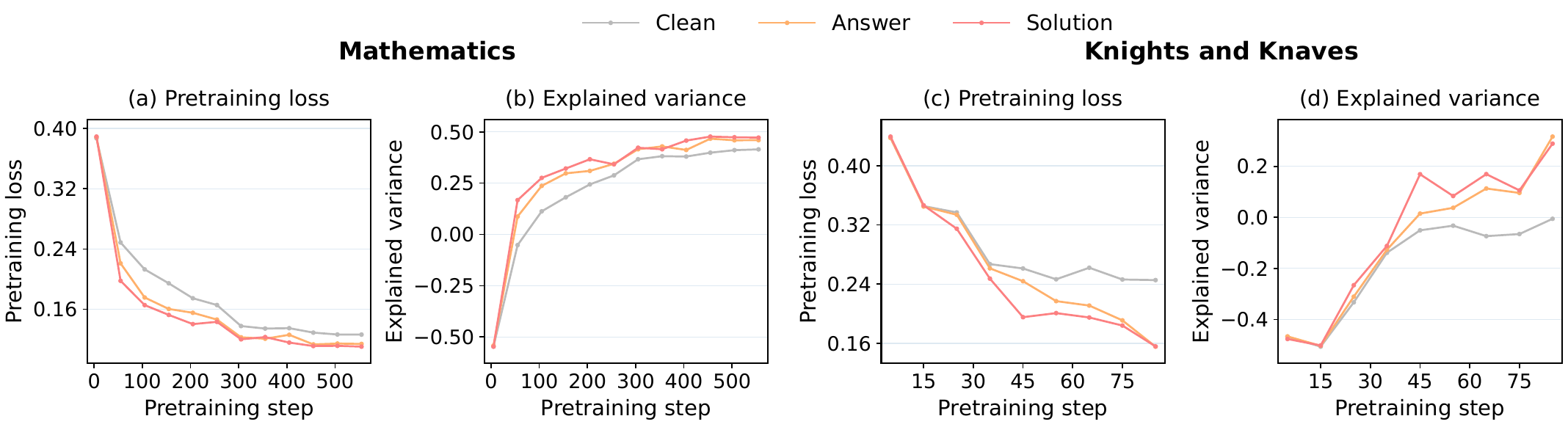}
    \caption{\textbf{Critic pretraining dynamics.}
    Logged loss and explained variance for clean, answer-conditioned,
    and solution-conditioned critics. Panels (a)--(b): mathematics, with
    12 recorded checkpoints from step 5 to 555. Panels (c)--(d): KK, with
    nine recorded checkpoints from step 5 to 85. Points show recorded values, and solid lines connect
    consecutive observations without additional smoothing.}
    \label{fig:critic-pretraining}
\end{figure*}

\subsection{Combining Distillation with Verifiable Rewards}

Several methods combine dense distillation with outcome optimization through
an additive objective or advantage.  MiMo-V2-Flash adds teacher--student
log-probability gaps to outcome-derived advantages in its multi-teacher
distillation pipeline \citep{xiao2026mimo}.  SDPG combines a verifier-based
policy-gradient objective with an exact full-vocabulary self-distillation
loss and reference-policy KL regularization \citep{liu2026selfdistilledpolicygradient}.
SAF-OPD studies the relative magnitude and training schedule of the two
signals, using sparsification and bounded compression together with warm-up
and annealing for the external-teacher OPD branch \citep{ding2026safopd}.
These designs show that the granularity, scale, and timing of teacher
supervision are separate choices in the fusion interface.

Other methods select where to apply each signal.  Sample-Routed Policy
Optimization (SRPO) routes failed responses with available teacher
information to self-distillation and the remaining responses to GRPO \citep{li2026unifying}.
Reward-Gated OPD instead retains trajectories for distillation according to
agreement between the verifier-derived advantage and a sequence-level
teacher--student likelihood gap \citep{akhondzadeh2026rewardgated}.  Its verifier thus
controls a distillation gate, rather than contributing an additive reward
term.  VASPO uses a learned prefix value to construct its advantage and
applies teacher modulation within that actor--critic update.

The closest group-relative interfaces use the teacher to refine
verifier-grounded credit. Self-Distilled RLVR (RLSD) preserves the sign of the
outcome advantage while a detached teacher--student evidence ratio changes
token-level update magnitudes \citep{yang2026selfdistilledrlvr}. RLCSD
contrasts correct and incorrect privileged contexts to suppress
correctness-independent style shifts in the teacher signal
\citep{pan2026rlcsdreinforcementlearningcontrastive}. SSPO applies
teacher--student disagreement as step-level advantage weights on incorrect
deep-search trajectories \citep{wu2026step}. Their group-relative outcome
advantages are shared across positions before teacher modulation; they do not
learn a value conditioned on privileged information for each student prefix.
VASPO uses the privileged information in a learned PPO critic as well as the
teacher, so its outcome-grounded advantages can vary across prefixes before
modulation. The bounded fusion preserves the sign of the raw privileged-critic
advantage (Section~\ref{sec:advantage-interface}).

\section{Critic Pretraining}
\label{app:critic-pretraining}

\paragraph{Pretraining Data.}
For each task and policy backbone, we use the initial policy checkpoint
to sample eight responses per training prompt. The actor observes only the
standard problem prompt, without the privileged solution or answer.
Each response receives a binary correctness label from the task verifier
described in Appendix~\ref{app:evaluation-details}. We upsample incorrect
response trajectories to obtain an approximately balanced ratio of correct
and incorrect examples. The same sampled trajectories and return labels
are used to pretrain the clean and privileged critic variants.

\paragraph{Pretraining Protocol.}
Each critic is initialized from its corresponding policy backbone with a
scalar value head and pretrained for two epochs before RL, following the
value-pretraining protocol of VAPO~\citep{yue2025vapo}. Pretraining minimizes
the mean squared error between each token-level prediction and the
response's terminal verifier return. With $\gamma=1$, every valid
response-token state has the same binary target. The clean critic reads
the prompt and preceding response tokens; the privileged critic additionally
reads the fixed dataset solution and answer. Neither critic observes the
current token or subsequent response tokens. PPO and the critic-based
reader ablations use the same initialization, pretraining trajectories,
targets, and epoch count; the input conditioning changes with the ablation.
Offline pretraining initializes the critic, while the per-rollout
pre-update in Appendix~\ref{app:implementation} adapts it during RL.

\paragraph{Pretraining Dynamics.}
Figure~\ref{fig:critic-pretraining} shows the recorded loss and explained
variance under clean, answer, and solution conditioning. In mathematics,
loss decreases overall and explained variance increases for all three
variants. At the last recorded checkpoint in each task, the answer- and
solution-conditioned critics have lower loss and higher explained variance
than the clean critic.

\section{Additional Experimental Details}
\label{app:experimental-details}

\subsection{Datasets and Models}
\label{app:datasets}

Mathematical reasoning uses Qwen3-4B-Instruct-2507 and
OLMo3-7B-Instruct-DPO, while logical reasoning uses
Qwen3-4B-Instruct-2507. We train mathematical models on the full
DeepMath-103K dataset~\citep{he2025deepmath}, using all 103,022 problems
without filtering or reserving a development subset from this dataset.
Logical reasoning uses the complete 5,000-problem training split of
Knights and Knaves (KK) puzzles with 4--8 roles~\citep{xie2025logic}. The solutions and
answers provided by each dataset form the privileged information shared
by the teacher and critic. All compared methods use the same task-specific
training data within each backbone.

We evaluate mathematical reasoning on AIME 2024, AIME 2025, AIME 2026,
HMMT 2025, Minerva Math, and OlympiadBench. OlympiadBench evaluation
uses all 674 problems, with none skipped. KK evaluation uses 500 held-out
4--8-role puzzles in total and 100 out-of-distribution puzzles for each
of the 9-, 10-, and 11-role settings. Table~\ref{tab:dataset-assets}
lists the dataset and model sources, licenses, and available question counts.

Qwen uses its native non-thinking Instruct template, and OLMo its native
instruction template.

\begin{table*}[t]
    \centering
    \footnotesize
    \setlength{\tabcolsep}{3pt}
    \renewcommand{\arraystretch}{1.12}
    \begin{tabular}{@{}lp{0.21\textwidth}p{0.40\textwidth}p{0.16\textwidth}r@{}}
        \toprule
        \textbf{Category} & \textbf{Asset} & \textbf{URL}
        & \textbf{License} & \textbf{Questions} \\
        \midrule
        Dataset & DeepMath-103K
        & \url{https://huggingface.co/datasets/zwhe99/DeepMath-103K}
        & MIT & 103,022 \\
        Dataset & Knights and Knaves (KK)
        & \url{https://github.com/Unakar/Logic-RL}
        & CC BY-NC-SA 4.0 & 6900 \\
        Dataset & AIME 2024
        & \url{https://huggingface.co/datasets/math-ai/aime24}
        & Apache-2.0 & 30 \\
        Dataset & AIME 2025
        & \url{https://huggingface.co/datasets/math-ai/aime25}
        & Apache-2.0 & 30 \\
        Dataset & AIME 2026
        & \url{https://huggingface.co/datasets/math-ai/aime26}
        & Apache-2.0 & 30 \\
        Dataset & HMMT 2025
        & \url{https://huggingface.co/datasets/FlagEval/HMMT_2025}
        & CC BY-NC-SA 4.0 & 30 \\
        Dataset & Minerva Math
        & \url{https://huggingface.co/datasets/math-ai/minervamath}
        & Not specified & 272 \\
        Dataset & OlympiadBench
        & \url{https://huggingface.co/datasets/math-ai/olympiadbench}
        & Not specified & 674 \\
        \midrule
        Model & Qwen3-4B-Instruct-2507
        & \url{https://huggingface.co/Qwen/Qwen3-4B-Instruct-2507}
        & Apache-2.0 & -- \\
        Model & OLMo3-7B-Instruct-DPO
        & \url{https://huggingface.co/allenai/Olmo-3-7B-Instruct-DPO}
        & Apache-2.0 & -- \\
        Model & CompassVerifier-3B
        & \url{https://huggingface.co/opencompass/CompassVerifier-3B}
        & Apache-2.0 & -- \\
        \bottomrule
    \end{tabular}
    \caption{Dataset and model assets. Licenses and question counts describe
    the linked distributions. ``Not specified'' means that the
    linked distribution does not state a license.}
    \label{tab:dataset-assets}
\end{table*}

\subsection{Evaluation Details}
\label{app:evaluation-details}

\paragraph{Training Stage.}
For mathematical reasoning, we use \texttt{math-verify}\footnote{
\url{https://github.com/huggingface/Math-Verify}} to compare each generated
answer with its reference answer, producing a binary terminal reward in
$\{0,1\}$. This supports the non-integer answers present in the full
DeepMath training set. For KK, the task verifier checks the complete
knight/knave assignment against the reference assignment; a response is
correct only when every role is correct. These verifier outcomes also
provide the critic-pretraining labels and RL return targets.

\paragraph{Test Stage.}
For all six mathematical benchmarks, we use a two-stage verification
procedure. We first evaluate every predicted answer with
\texttt{math-verify}. Responses judged correct are accepted
directly, while those judged incorrect are sent to
\texttt{CompassVerifier-3B} for a second verification pass. For these
responses, the second-stage judgment determines the final correctness
label. KK evaluation uses the complete-assignment verifier described above.

\paragraph{Sampling and Metrics.}
Evaluation uses temperature $0.6$, top-$p$ $0.95$, top-$k$ $20$, and a
maximum response length of $8{,}192$ tokens for both backbones. We use
$K=32$ samples per problem for AIME24, AIME25, AIME26, and HMMT25,
$K=8$ for Minerva Math and OlympiadBench, and $K=16$ for KK.
For a benchmark with $N$ problems, the reported percentage is
\begin{equation}
    \operatorname{avg@}K
    =\frac{100}{KN}\sum_{j=1}^{N}\sum_{k=1}^{K}
      \mathcal{R}(x_j,y_{j,k}).
    \label{eq:math-avg}
\end{equation}
This metric estimates single-sample correctness from repeated draws.
Math Average weights the six benchmarks equally; KK Average weights the
in-distribution score and the three out-of-distribution scores equally,
despite their different problem counts. All methods use the same decoding
and verification procedure within each backbone. Trained methods use the
last checkpoint at the fixed rollout budget; test scores are excluded
from checkpoint and hyperparameter selection. Problem-bootstrap intervals
keep all responses from a problem together.

\paragraph{Critic Explained Variance.}
EV is defined in \eqref{eq:critic-ev}. Controlled critic comparisons score
the same clean-student responses at the same valid token positions.
Groups with zero return variance are reported as undefined.

\subsection{Baselines}
\label{app:baselines}

We compare VASPO with three groups of baselines. Vanilla reports the initial
policy checkpoint without RL. DAPO is the critic-free RLVR baseline, while
PPO uses a clean critic and the same value-pretraining protocol as VASPO.
OPSD~\citep{zhao2026opsd} and RLSD~\citep{yang2026selfdistilledrlvr}
compare against privileged self-distillation and its combination with
verifiable rewards. External baselines retain their documented objectives.

For the reader-component comparisons, the variant without the privileged
critic removes the entire privileged critic and its associated PPO/GAE
update path, and trains with the OPSD objective. The variant without
self-teacher modulation retains the privileged critic and PPO update,
while removing both reader components yields PPO with a clean critic.
PPO, Critic-only, and VASPO share critic architecture, initialization, pretraining,
value targets, data order, and optimization settings within each seed.
These three conditions also share
normalization, critic pre-update schedule, rollout correction,
reference-policy KL settings, and actor loss reduction.
Each condition generates its own on-policy responses.

\subsection{Training Details}
\label{app:training-details}

\paragraph{Optimization Settings.}
We implement VASPO in \texttt{verl}~\citep{sheng2024hybridflow} and update
all model parameters with AdamW. The global batch size is 128, the PPO
mini-batch size is 32, and the actor and critic learning rates are
$1\times10^{-6}$ and $5\times10^{-6}$, respectively. We set
$\gamma=\lambda_{\mathrm{GAE}}=1$ and use PPO clipping
$\epsilon_{\mathrm{clip}}=0.2$. VASPO uses full clipped probability-ratio
reweighting with $\epsilon_w=0.2$ and token-selection parameters
$\tau=\delta=0.02$, without advantage normalization. The self-teacher
uses EMA decay $0.95$ after each actor optimizer step.

\paragraph{Rollout Settings.}
We sample eight responses per prompt at temperature $1.0$, top-$p$ $0.95$,
and top-$k$ $20$. The maximum response length is $8{,}192$ tokens for
both backbones. The clean actor prompt limit is $2{,}048$ tokens,
including its chat template. Privileged reader inputs have a separate
context budget: the complete solution and answer, reader wrapper, problem,
and response prefix must fit the reader's context window. Oversized or
invalid privileged inputs are handled before the common training manifest
is frozen, with excluded counts recorded. Evaluation retains every test
problem and logs token-limit termination and answer-extraction failures.
Prompt templates are listed in Appendix~\ref{app:prompts}.

\paragraph{Hyperparameter Record.}
Table~\ref{tab:planned-protocol} summarizes training and decoding settings;
Table~\ref{tab:reader-configuration} records reader and update settings.
The working configuration includes 50-step linear warmup followed by a
constant learning rate, weight decay $0.01$, one PPO epoch per batch,
and a budget of 200 rollout-batch updates. We set the reference-policy
KL coefficient to $0$. The launcher defaults include no entropy bonus,
value-loss coefficient $0.5$, and token-level rollout importance weighting.

\begin{table}[t]
    \centering
    \small
    \setlength{\tabcolsep}{4pt}
    \renewcommand{\arraystretch}{1.08}
    \begin{tabular}{@{}p{0.57\columnwidth}p{0.36\columnwidth}@{}}
        \toprule
        \textbf{Parameter} & \textbf{Value} \\
        \midrule
        Training framework & \texttt{verl} \\
        Training hardware & $8\times$ H200 GPUs \\
        Parameter updates & Full-parameter \\
        Optimizer & AdamW \\
        Weight decay & $0.01$ \\
        Critic pretraining epochs & 2 \\
        Actor learning rate & $1\times10^{-6}$ \\
        Critic learning rate & $5\times10^{-6}$ \\
        Learning-rate schedule & constant \\
        Global batch size & 128 \\
        PPO mini-batch size & 32 \\
        Full-run rollout updates & 200 \\
        PPO epochs per batch & 1 \\
        Discount factor $\gamma$ & $1.0$ \\
        GAE parameter $\lambda_{\mathrm{GAE}}$ & $1.0$ \\
        PPO clipping $\epsilon_{\mathrm{clip}}$ & $0.2$ \\
        PPO clipping (higher) & $0.28$ \\
        Value-loss coefficient $c_V$ & $0.5$ \\
        Reference-policy KL coefficient & $0$ \\
        Entropy-loss coefficient & $0$ \\
        Responses per prompt & 8 \\
        Rollout / evaluation temperature & $1.0$ / $0.6$ \\
        Top-$p$ / top-$k$ & $0.95$ / $20$ \\
        Maximum response length & $8{,}192$ tokens \\
        Clean actor prompt limit & $2{,}048$ tokens \\
        Main checkpoint & Last at fixed rollout budget \\
        \bottomrule
    \end{tabular}
    \caption{Training and decoding configuration for VASPO and matched
    reader ablations.}
    \label{tab:planned-protocol}
\end{table}

\subsection{Training-time Analysis}
\label{app:training-time}

\begin{figure*}[t]
    \centering
    \includegraphics[width=\textwidth]{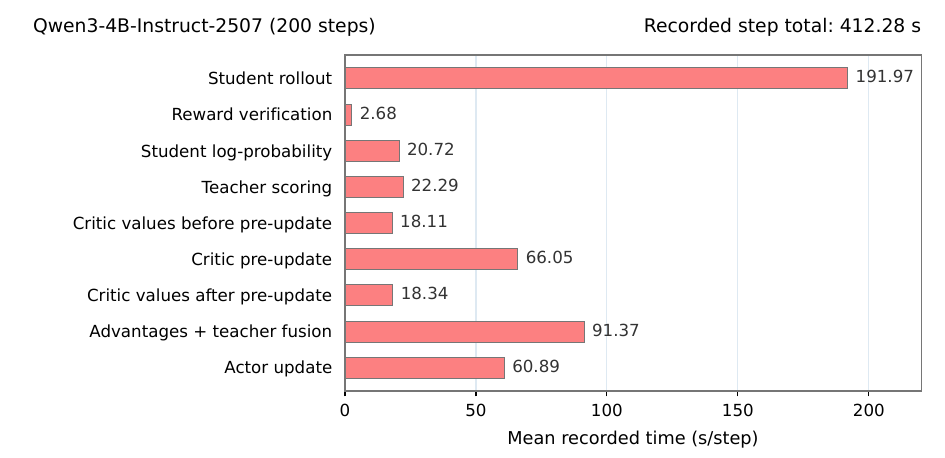}
    \caption{\textbf{Training-time profile of VASPO with Qwen3-4B-Instruct-2507.}
    Mean recorded time in seconds per step over 200 training steps.
    Bars show individual component timers; the separately recorded
    end-to-end step time is $412.28$\,s. The component timers are not
    an additive decomposition of this total.}
    \label{fig:training-time}
\end{figure*}

Figure~\ref{fig:training-time} reports mean timer values for
Qwen3-4B-Instruct-2507 over 200 training steps. The recorded end-to-end
step time is $412.28$\,s. Student rollout
has the largest component timer at $191.97$\,s, while reward verification
and student log-probability computation take $2.68$\,s and $20.72$\,s,
respectively. The privileged self-teacher scores the sampled student
tokens in $22.29$\,s per step.

The critic schedule includes value inference both before and after its
pre-update, taking $18.11$\,s and $18.34$\,s, respectively; the critic
update takes $66.05$\,s. Actor training takes $60.89$\,s. The timer for
advantage computation and teacher fusion records $91.37$\,s, measuring
these operations jointly rather than isolating the fusion overhead.

The component timer means sum to $492.42$\,s, exceeding the separately
recorded step time. We therefore report the timers individually rather
than treating them as an additive partition of end-to-end time.

\subsection{Prompts}
\label{app:prompts}

We list the actor instruction and the two privileged-input templates for
each reader. Braced fields denote per-example substitutions:
\texttt{\{problem\}} is the problem text, \texttt{\{answer\}} is the
reference final answer, and \texttt{\{solution\}} is the reference worked
solution. In the critic templates, \texttt{\{actor user message\}} denotes the
complete original actor user message, including any rollout instruction.
The boxes show message content before the tokenizer's native chat template
is applied. Line wrapping is for readability; paragraph breaks follow the
templates.

\paragraph{Actor instruction.}
The actor receives the problem without privileged information. The
data-preprocessing script places the following instruction in a system
message, followed by the problem in a user message.

\begin{promptbox}{gray}{Actor: system instruction}
Please reason step by step, and put your final answer within
\textbackslash boxed\{\}.
\end{promptbox}

The self-distillation rollout formatter instead includes the same
instruction in a single user message: \texttt{Problem: \{problem\}}, a
blank line, and the instruction above for training; its evaluation format
omits the \texttt{Problem:} header. Thus, the instruction text is shared,
while its message role depends on the input-construction path.

\paragraph{Self-teacher: solution and answer.}
The solution-and-answer template uses the following user message.
The worked solution precedes the explicit final answer, and both lie
inside the reference-solution block.
This template adds no system message.

\begin{promptbox}{blue}{Self-teacher: solution and answer}
Problem: \{problem\}

\medskip
Here is a reference solution to this problem:\\
=== Reference Solution Begin ===\\
\{solution\}

\medskip
Correct final answer: \{answer\}\\
=== Reference Solution End ===

\medskip
After reading the reference solution above, make sure you understand
the reasoning behind each step.

\medskip
Please reason step by step, and put your final answer within
\textbackslash boxed\{\}.
\end{promptbox}

\paragraph{Self-teacher: answer only.}
The answer-only template uses the following user message. It
provides the final answer as a verified target and asks the teacher to
preserve valid reasoning from the current prefix. This template also adds
no system message.

\begin{promptbox}{blue}{Self-teacher: answer only}
Problem: \{problem\}

\medskip
=== Verified Target Begin ===\\
Final answer: \{answer\}\\
=== Verified Target End ===

\medskip
Use the verified target only to check the correctness of the reasoning.\\
The target does not specify a unique solution path.\\
Continue from the current reasoning prefix and preserve any valid approach.\\
Do not mention, quote, or refer to the verified target or its existence.

\medskip
Please reason step by step, and put your final answer within
\textbackslash boxed\{\}.
\end{promptbox}

\paragraph{Critic: answer only.}
The privileged critic preserves the original actor messages, including
any system message, and appends the following answer suffix to the last
user message. A single space separates the original user content and
the suffix. The clean critic uses the original actor messages without
this addition.

\begin{promptbox}{orange}{Critic: answer only (user message)}
\{actor user message\} The ground truth answer is \{answer\}.
\end{promptbox}

\paragraph{Critic: solution and answer.}
When solution conditioning is enabled, the critic additionally appends a
blank line and the reference solution after the answer suffix. The
answer therefore precedes the solution in the critic input.

\begin{promptbox}{orange}{Critic: solution and answer (user message)}
\{actor user message\} The ground truth answer is \{answer\}.

\medskip
A reference solution:\\
\{solution\}
\end{promptbox}

After chat-template formatting, both privileged readers are evaluated on
the actor's sampled response tokens. The self-teacher scores those tokens,
and the critic estimates prefix values; neither reader generates a separate
response in these scoring passes.

\section{Additional Method Details}
\label{app:method}

\paragraph{Value regression and advantage computation.}
The critic regresses to student-rollout returns using
\eqref{eq:privileged-value-loss}. The training driver first evaluates values
and constructs return targets. These initial predictions remain the fixed
clipping centers during the critic pre-update. The driver then re-evaluates
values and recomputes the actor advantages. Under terminal-only rewards and
$\gamma=\lambda_{\mathrm{GAE}}=1$, the target is $G_i$ at every valid token,
and the refreshed advantage is $G_i-v^{z,+}_{i,t}$ as in
\eqref{eq:privileged-advantage}. Only these recomputed advantages enter token
selection and teacher fusion. Their critic snapshot is $\phi^+$, with
terminal value $V^z_{\phi^+}(h_{T+1},z)=0$. The privileged
information remains fixed throughout the response, and the critic's input
excludes the current token and subsequent continuation. When
$z=f(x)$ is deterministic, conditioning on privileged information changes the critic's
representation without adding information beyond the prompt; the question is
whether this explicit representation helps a finite critic predict student
returns.

\paragraph{Effect of reweighting.}
For $A^z_t\ne0$, \eqref{eq:direction-preserving-fusion} satisfies
\begin{equation}
    1-\epsilon_w\leq\frac{A^{z,\mathrm{f}}_t}{A^z_t}\leq1+\epsilon_w.
    \label{eq:fusion-bound}
\end{equation}
Selected tokens receive the clipped ratio, while unselected tokens have
unit weight. This bound preserves the sign of the raw privileged-critic
advantage. Changing the critic can still change that advantage relative to
a critic without privileged input. Algorithm~\ref{alg:shared-privilege}
summarizes the signal flow; the teacher-update schedule and auxiliary
training settings are specified in Appendix~\ref{app:implementation}.

\section{Complete Implementation Details}
\label{app:implementation}

This section records the experimental realization of the method, including
teacher updates, token selection, loss reduction, and optional variants.

\begin{table*}[t]
    \centering
    \small
    \setlength{\tabcolsep}{5pt}
    \renewcommand{\arraystretch}{1.08}
    \begin{tabular}{@{}p{0.35\textwidth}p{0.60\textwidth}@{}}
        \toprule
        \textbf{Setting} & \textbf{Value or procedure} \\
        \midrule
        Critic architecture & Actor backbone with a scalar value head;
        matched initialization across clean and privileged variants \\
        Critic regression & Monte Carlo terminal return;
        PPO clipped value loss \\
        Critic update schedule & Pre-update once per rollout batch,
        then re-score values and recompute GAE before fusion \\
        Fusion & Full clipped probability-ratio reweighting;
        $\epsilon_w=0.2$ \\
        Token selection & Raw $A^z_t>0$ and
        $|\tanh(d_t/\tau)|>\delta$; $\tau=\delta=0.02$ \\
        Student probability source & Behavior policy \\
        Advantage normalization & None before or after fusion \\
        Teacher update & EMA decay $\beta_{\mathrm{EMA}}=0.95$
        after each actor optimizer step \\
        Teacher score caching & Once per rollout batch, before actor updates \\
        Actor loss reduction & Mean over valid tokens per response,
        then over responses \\
        Rollout correction & Token-level importance weighting
        (launcher default) \\
        \bottomrule
    \end{tabular}
    \caption{Critic, teacher, and policy-update settings for VASPO. Matched controls
    retain the applicable settings when a reader is disabled.
    Optional implementation variants are described in
    Appendix~\ref{app:implementation}.}
    \label{tab:reader-configuration}
\end{table*}

\paragraph{Critic update schedule.}
The critic is updated before computing the actor advantages for each rollout
batch. We first construct the critic's return targets and store its value
predictions as fixed clipping centers, run the PPO clipped value update,
and repeat value inference and GAE on the same rollout batch.
Teacher fusion is applied only after this second pass. The ordinary later
critic-update call is skipped, so the schedule changes when the critic is
trained without adding a second update phase. This is a per-batch operation,
distinct from the offline initialization in
Appendix~\ref{app:critic-pretraining}.

\paragraph{Teacher parameters and score caching.}
The self-teacher uses an exponential moving average of student parameters,
initialized with $\bar\theta_0=\theta_0$. After each actor optimizer step,
\begin{equation}
    \bar\theta_{k+1}
    =\beta_{\mathrm{EMA}}\bar\theta_k
     +(1-\beta_{\mathrm{EMA}})\theta_{k+1}.
    \label{eq:teacher-ema}
\end{equation}
The decay is listed in Table~\ref{tab:reader-configuration}. Teacher
probabilities are evaluated and cached once per rollout batch, before the
actor update. They remain fixed within that batch even as the EMA
parameters change.

\paragraph{Token selection and reweighting.}
The implementation computes the detached gap in \eqref{eq:teacher-gap}
and selects valid response tokens with \eqref{eq:teacher-token-mask}.
Table~\ref{tab:reader-configuration} gives the temperature, threshold, and
clipping width. The $\tanh$ transform only selects tokens; their weight
is the clipped ratio $e^{d_t}$ from \eqref{eq:teacher-weight}.
Selected advantages are multiplied directly by this clipped ratio.
Negative and zero raw advantages,
and positive advantages below the signal threshold, retain their GAE values.

\paragraph{Behavior versus current student probabilities.}
The main formulation uses $p_{\mathrm{old},t}(y_t)$, keeping the weights,
selection mask, and fused advantages fixed during the batch update.
A supported variant replaces this denominator with $p_{\theta,t}(y_t)$
and recomputes the gap, mask, and fusion at each actor forward pass.
Teacher probabilities remain cached in either case. The recomputed fusion
is still detached, so gradients do not pass through its denominator.
PPO's update ratio always compares the current actor with the behavior
policy. Recomputing teacher scores during actor updates is a separate
option, disabled in the default configuration.

\paragraph{Advantage normalization.}
The default uses raw GAE advantages for selection and fusion, with no
whitening before or after fusion. If post-fusion whitening is enabled,
let $\mathcal{M}$ contain every valid token position in the rollout batch:
\begin{equation}
    \widehat A_{i,t}
    =\frac{A^{z,\mathrm{f}}_{i,t}-\mu_{\mathcal{M}}}
           {\sqrt{\sigma^2_{\mathcal{M}}+10^{-8}}}.
    \label{eq:whitened-fused-advantage}
\end{equation}
The mean and variance include selected and unselected tokens. Both loss
paths use these same statistics, and the selection mask is not recomputed.
Centering can change advantage signs, so \eqref{eq:fusion-bound} describes
the fusion before whitening. With raw GAE input, this normalization option
requires the behavior-policy denominator and cached teacher scores; the
current-policy variant uses no post-fusion whitening.

\paragraph{Loss reduction and auxiliary settings.}
Let $B$ be the number of responses and $T_i$ their valid-token counts.
Write $\widehat A_{i,t}=A^{z,\mathrm{f}}_{i,t}$ by default, or use
\eqref{eq:whitened-fused-advantage} when whitening is enabled. The core
actor loss averages tokens within each response and then averages responses:
\begin{equation}
    \begin{aligned}
        &\mathcal{L}_{\mathrm{actor}}(\theta)
        =-\frac{1}{B}\sum_{i=1}^{B}\frac{1}{T_i}\sum_{t=1}^{T_i}\\
        &\quad\min\!\left(
            \rho_{i,t}(\theta)\widehat A_{i,t},
            \rho^{\mathrm{clip}}_{i,t}(\theta)\widehat A_{i,t}
        \right).
    \end{aligned}
    \label{eq:vaspo-actor-loss}
\end{equation}
Selected and unselected contributions share the valid-token denominator
and have equal branch weights. The PPO ratio and its clipping are defined
in Section~\ref{sec:rlvr}; the probability-ratio clip $\epsilon_w$ instead
bounds advantage reweighting. The launcher additionally applies token-level
rollout importance weights to this surrogate. The reference-policy KL
coefficient is $0$, as listed in Table~\ref{tab:planned-protocol}.
The critic uses the regression loss in \eqref{eq:privileged-value-loss}.

\section{Additional Experiments}
\label{app:experiments}

We trace the positional pattern in Section~\ref{sec:position-motivation}
from saved responses to independent continuations, then ask which response
groups account for it. The evidence identifies where critic predictions
improve; teacher utility and training interaction require separate tests.

\subsection{Signal Profiles on Fixed Responses}
\label{app:frozen-response-position}

\begin{figure*}[t]
    \centering
    \includegraphics[width=\textwidth]{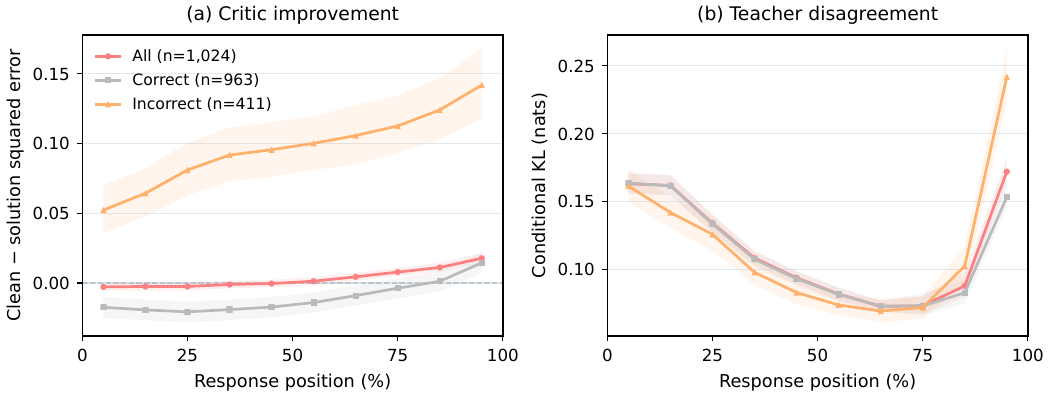}
    \caption{\textbf{A pooled trend hides different response groups.}
    Saved-response critic error reduction (left) and teacher KL (right).
    Shading shows pointwise 95\% problem-bootstrap intervals; $n$ counts
    problems, which can occur in both correctness groups.
    There are 6,659 correct and 1,629 incorrect responses (8,288 total).
    Positive critic values favor solution conditioning; teacher KL measures
    disagreement, not useful guidance.}
    \label{fig:frozen-response-groups}
\end{figure*}

To separate the pooled trend from its contributing groups, we score ordinary
and solution-conditioned critics and the self-teacher on identical saved
response tokens. The 1,024 problems contribute 8,288 responses, split by
correctness; one problem can appear in both groups.
With realized outcome $R$, the critic signal is
\begin{equation}
    \Delta_t^V=(v_t^0-R)^2-(v_t^z-R)^2,
    \label{eq:frozen-response-value-improvement}
\end{equation}
and the teacher signal is conditional $D_{\mathrm{KL}}(p_{\theta,t}\|q^z_{\theta,t})$.
Values are read before the current token. Tokens are assigned to ten bins
using $(t-\tfrac12)/T$ without rescaling after exclusions.

Figure~\ref{fig:frozen-response-groups} reveals two different critic
profiles: errors decrease throughout incorrect responses, whereas correct
responses incur an early cost and benefit only near the end. Teacher KL
falls toward the middle and rebounds at the end in both groups. Thus,
the pooled curve hides outcome-dependent critic behavior, and teacher
change is not confined to the beginning. Since each response supplies only
one realized outcome, we next check critic predictions against independently
sampled continuations.

\subsection{Prediction Quality under Independent Continuations}
\label{app:value-diagnostic}

\begin{figure*}[t]
    \centering
    \includegraphics[width=\textwidth]{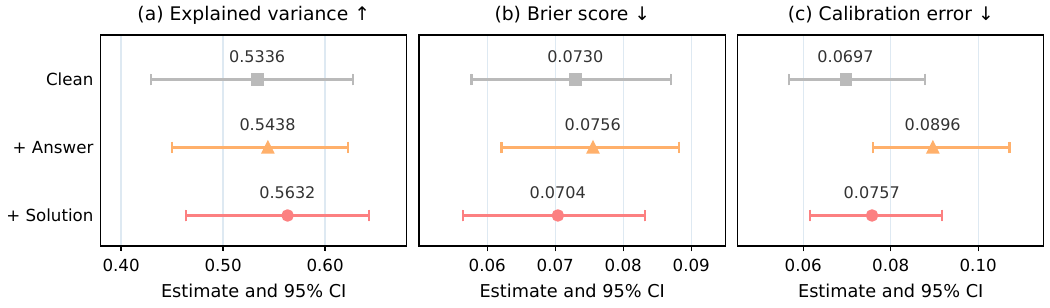}
    \caption{\textbf{Prediction accuracy and calibration need not improve together.}
    Three existing critics evaluated on the same 128 problems, 384 prefixes,
    and 3,072 independent continuations. Markers and bars show estimates and
    pointwise 95\% problem-bootstrap intervals. Suffix labels agree with
    full-response rescoring for all 3,072 continuations. Arrows indicate
    the better direction.}
    \label{fig:value-diagnostics}
\end{figure*}

We freeze Qwen3-4B-Instruct-2507 and compare three existing critics given
the clean prefix, the prefix with an answer, or the prefix with an answer
and first reference solution. On 128 evaluation problems, we sample eight
clean-actor continuations from each of three prefixes at relative positions
$0.2$, $0.5$, and $0.8$. All critics predict the same continuation outcomes.

Figure~\ref{fig:value-diagnostics} shows the highest EV and lowest Brier
point estimates for solution conditioning, but the clean critic has the
lowest ECE. Better prediction error therefore does not imply better
calibration. These point-estimate rankings alone do not establish a paired
overall advantage. Full-response rescoring confirms every evaluation label
(Appendix~\ref{app:diagnostic-protocol}).

\subsection{Where the Critic Improvement Occurs}
\label{app:position-analysis}

\begin{figure*}[t]
    \centering
    \includegraphics[width=\textwidth]{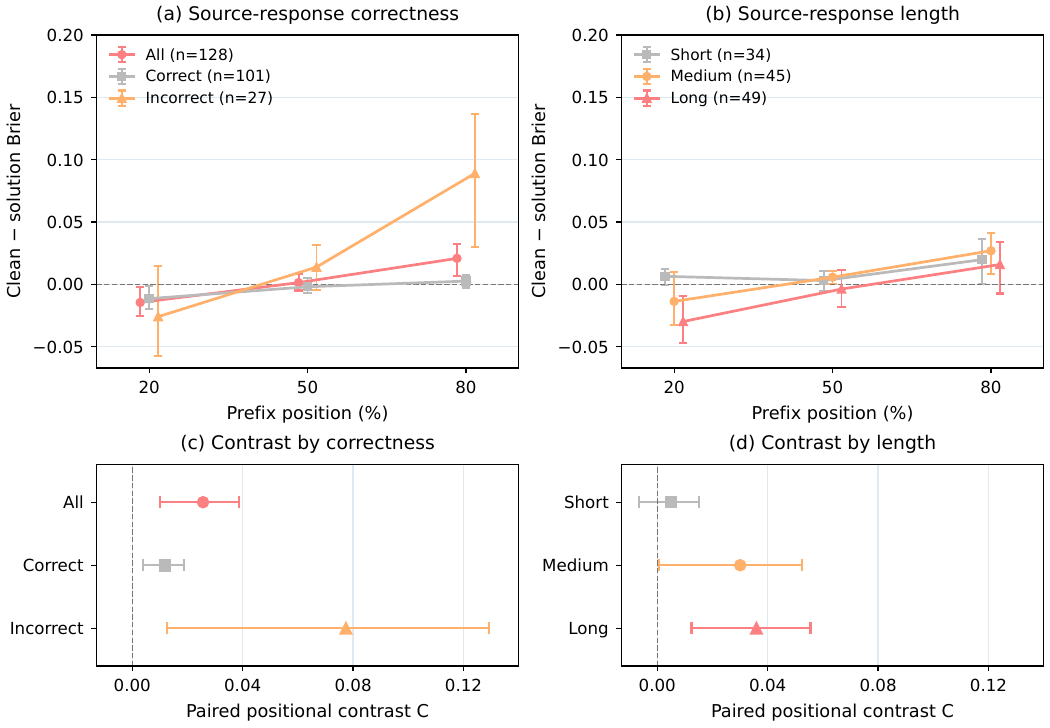}
    \caption{\textbf{Later critic gains depend on the source response.}
    Clean-minus-solution Brier improvement by source correctness (a) and
    length (b), with paired positional contrasts $C$ in (c,d).
    Positive values favor the solution critic. Bars are pointwise 95\%
    problem-bootstrap intervals; $n$ counts problems. Horizontal offsets
    in (a,b) separate overlapping points; the sampled positions are 20\%,
    50\%, and 80\%. Each row shares axis limits. Length strata use fit-split
    tertiles (1,008 and 1,779.33 tokens, upper bounds included).
    All evaluation labels agree with full-response rescoring.}
    \label{fig:value-position-diagnostics}
\end{figure*}

Aggregate errors can hide offsetting gains and costs. We therefore compare
clean-minus-solution Brier improvement $\Delta(r)$ at each sampled position
$r$, then form the within-problem contrast
\begin{equation}
    C=\frac{\Delta(0.5)+\Delta(0.8)}{2}-\Delta(0.2).
\end{equation}
This compares three discrete positions, not continuous trajectory regions.

Figure~\ref{fig:value-position-diagnostics}(a,c) shows an early cost and a
late benefit in the pooled group, with an uncertain middle-position effect.
The late gain is much larger for incorrect source responses; the correct
source group's late interval includes zero. Both groups have positive
positional contrasts. The relevant distinction is thus where prediction
improves within a response, rather than a uniform gain at every prefix.

Length controls refine this finding (Figure~\ref{fig:value-position-diagnostics}b,d).
The positional contrast is supported in medium and long responses but
uncertain in short ones. Yet the long-response late-gain interval includes
zero: its positive contrast can also reflect the pronounced early cost.
A positive positional contrast therefore does not establish a significant
late gain in every length stratum. Final-answer and terminal-region
exclusions are unavailable because these regions were not marked.

\subsection{Diagnostic Protocol}
\label{app:diagnostic-protocol}

\paragraph{Samples and models.}
The continuation study contains 256 problems (768 prefixes, 6,144 continuations).
A separate 128-problem fit split supplies length-tertile thresholds, computed
by linearly interpolated percentiles. Short responses have $T\leq1{,}008$;
medium responses have $1{,}008<T\leq1{,}779.333\ldots$; longer responses
form the long stratum. The critics are not refitted on this split.
The evaluation split has 101 correct and 27 incorrect source responses,
and 34 short, 45 medium, and 49 long responses. Its 3,072 continuation
returns contain 2,517 successes and 555 failures.
Each scalar-headed Qwen3 critic was trained on 577,140 balanced examples
from the DeepMath training source for two epochs, at learning rate
$5\times10^{-6}$, with 256-step gradient accumulation, bfloat16, and
ZeRO-1 on eight GPUs.

\paragraph{Targets and metrics.}
The verifier compares the last boxed answer with the ground truth. Original
labels scored the new suffix; rescoring all 3,072 evaluation continuations
after prefix concatenation changes none of them (1,024 checked at each
position, none unrecoverable). Separate and joint token decoding agree,
including for the 17 prefixes that already contain boxed expressions.
There were no verifier execution failures. The evaluation split contains
96 truncated continuations, out of 192 across both splits.
For predictions $p_h$ and binary outcomes $y_{hk}$,
\begin{equation}
    \mathrm{Brier}_h=(p_h-\bar y_h)^2+
       \frac{1}{K_h}\sum_k(y_{hk}-\bar y_h)^2.
\end{equation}
The shared variance term cancels in paired critic differences, so we plot
Brier improvement without a redundant MSE-improvement panel.
EV uses pooled weighted moments,
\begin{equation}
    \mathrm{EV}=1-
    \frac{\mathbb{E}_w[(y-p)^2]-\mathbb{E}_w[y-p]^2}
         {\mathbb{E}_w[y^2]-\mathbb{E}_w[y]^2},
    \label{eq:diagnostic-weighted-ev}
\end{equation}
with NaN for return variance below $10^{-12}$; negative EV is retained.
ECE uses ten equal-width probability bins.

\paragraph{Aggregation and uncertainty.}
Both diagnostics weight problems equally. Within a problem, saved-response
curves average tokens within a response bin, then responses; continuation
curves average outcomes within a prefix, then prefixes. The pooled group
is recomputed from all records. Intervals use 2,000 whole-problem percentile
bootstrap replicates; continuation comparisons share sampled indices
(seed 20261010), recomputing EV and ECE each time. Intervals are pointwise,
without multiplicity correction. The original fixed-response protocol
retrieves the first solution by exact problem matching, uses the
solution-and-answer teacher template in Appendix~\ref{app:prompts}, and
excludes overlength inputs as complete paired records.

\end{document}